# Study of the Importance of Adequacy to Robot Verbal and Non Verbal Communication in Human-Robot Interaction

Céline Jost

*Lab-STICC*
*Vannes, France*

Brigitte Le Pévédic

*Lab-STICC*
*Vannes, France*

Dominique Duhaut

*Lab-STICC*
*Vannes, France*

*e-mail: celine.jost@univ-ubs.fr      brigitte.le-pevedic@univ-ubs.fr      dominique.duhaut@univ-ubs.fr*

*Abstract*—**The Robadom project aims at creating a homecare robot that help and assist people in their daily life, either in doing task for the human or in managing day organization. A robot could have this kind of role only if it is accepted by humans. Before thinking about the robot appearance, we decided to evaluate the importance of the relation between verbal and nonverbal communication during a human-robot interaction in order to determine the situation where the robot is accepted. We realized two experiments in order to study this acceptance. The first experiment studied the importance of having robot nonverbal behavior in relation of its verbal behavior. The second experiment studied the capability of a robot to provide a correct human-robot interaction.**

## I. INTRODUCTION

Communication is a complex domain. Everybody communicates whenever and everywhere. Communication is omnipresent but is difficult to define. Since sixty years, researchers have suggested definitions. Lasswell and Shannon gave basic components indicating that a transmitter sends a message, through a communication channel, until a receiver [1][2]. The message can be affected by interferences and disturbances. These first definitions have been followed by numerous models and schemes. They added precisions or new communication points of view. For example, Cherry indicated that a message depends on its content [3]. Riley added the importance of a group. A reciprocity phenomenon appeared showing that an individual communication is influenced by its group [4]. Picard specified that communication is a social process [5] which is multichannel according to Cosnier [6]. Signals can be sound; gestures; comical expressions; posture; chemical, tactile, thermal and electric exchanges. Gerbner explored a new way: a person which communicates ensures to have sent correct signals by controlling each sending [7]. It is an introduction to feedback. This explanation is simplistic and does not express each characteristic about communication: context, social conventions, emitter intention, cooperation between individuals, metacommunication... (defined by Blanchet [8]). In the human-robot interaction context, a simplistic communication definition is sufficient. Indeed, its purpose is to put the human in contact with the robot in order to obtain an "interunderstanding". According to

Grandgeorge, "interunderstanding" appears when communication signals are correctly perceived by the receiver and when the emitter is able to distinguish the good reception of his/her own message [9].

In a human-robot interaction context, communication between the both entities reveals an unknown data: robot communication. Everything has to be created because robots are empty boxes. Concerning robots expressions, Kirby decided that the robot behavior has to mimic human behavior because it allows to obtain a fluent and natural interaction [10]. Moreover, Gong and Kim consider that machines anthropomorphism increases positive human judgment and interaction quality [11][12]. Thus, robots have verbal communication which is speech and nonverbal communication which is gesture, posture, facial expressions... Concerning verbal communication, it is difficult to express emotions through speech because voice synthesis have technical handicaps. They are not able to correctly express prosody, tone, speed... as explained in [13]. Morris and Breazeal chose to add nonverbal communication in order to increase the dialog understanding with physical evidences [14][15][16]. The literature shows a lot of work which make a robot communicating. But, does the nonverbal communication really affect the human being which interacts? Ekman showed that no channel predominates in the human-human communication [17], but some studies underline that it is important to consider several channels to recognize a human emotion (see [18]). Is it the same process for a human which wants to decode robot's expression? Tojo gave an answer part in [19]: it seems that nonverbal behavior is important to precise a speech or to make it understandable. But, does a human really expect a robot to have a similar communication? The answer is a fundamental step toward "interunderstanding". Indeed, this one can only exist if human correctly perceives robot signals.

To answer these questions, two studied were carried out. The first one aimed at compare a robot credibility and sincerity according to its gesture. The speech was frozen. The problematic was to know whether a lack of adequacy between speech and gesture was able to affect the human perception. The second one aimed at study robot acceptability during an interaction with a human. Can some incoherence between a robot speech and gesture can affect its credibility and sincerity in the entire interaction? The answer of these questions allows to determine whether nonverbal communication is decisive for the human to understand robot.

## II. CONTEXT: THE ROBADOM PROJECT

The Robadom project [21], which is supported by the national research agency, aims at designing a home care robot which will daily assist the elderly. This robot has several roles:

(1) it will supervise and protect the patient, it is important to know whether the person is well and it is important to react whether the person has problems. But the protection begins with prevention. The robot can analyze what happens around it and must indicate if there is a danger. Moreover, it is a help for doctors when it reminds to patients to take their medicine.

(2) the robot is an assistant which manage the shopping lists, appointments etc.

(3) It is an entertainment because, as a companion, the robot can speak with the person, can play with the person etc.

(4) It is a social intermediary which can launch visual communication with the family, or give information about news etc.

The current patient are people with cognitive impairment, so the robot has to offer cognitive exercises. The project objective is to study the impact of such a robot on the elderly to know if it could be a solution to the ageing problems. The elderly is a concern in the entire world. For example, Heerink et al [22] tested the influence of a robot's social abilities on acceptance of elderly users but no correlation has be found between social abilities and technology acceptance. It is difficult to know what the elderly need exactly and what they want exactly. A study about game design for senior citizens [23] indicates that the elderly rejects computer because it can not replace a real person. It seems that these persons need to be useful, need to cultivate themselves and need to be connected to the society. The loneliness is the worst situation. That is why, the Robadom project wants to fight against the loneliness of the persons. Tefas et al [24] covered a part if this work by developing an application which can: (1) supervise the meal of the person and remind them to eat if they forget. (2) detect the facial expression of the person in order to analyze his/her emotion and express back an appropriate emotion.

The target population of the Robadom project are the most difficult because they did not grow up with technology, computing, robotics. They are often resistant to this project. **That is why the question of credibility, sincerity and acceptability is so important in this project. If the robot is not accepted, our patient will never interact with it.**

## III. STUDY 1: IMPORTANCE OF ADEQUACY BETWEEN SPEECH AND GESTURE

This study allowed to measure the importance of gesture according to a robot speech. The robot made three different gestures for a single sentence. An observer watched the interaction and gave her/his feeling after. This experiment used the same experimental protocol that [20] in their virtual agent studies.

Participants were 60 French people (17 women and 43 men) between 10 and 57 years old (mean: 18.1).

### A. Experimental design

The experiment was made at a robotic competition at an isolated stand. Participants were either competitors or spectators of the competition. Everyone of them had knowledge about robotic and an interest for robots. Each participant had to observe the robot and give her/his impression by completing a questionnaire.

### I. Equipment

A small humanoid robot (see Fig. 1) was placed on a table, face to participants. The experimenter was sitting on the next in order to control a computer which allow to the robot to move.

### II. Experimental setting and data recording

First of all, the experimenter asked participants to imagine the following context: the robot had advised them to watch a movie. Back home, two scenarios were possible. Scenario A: participants did not like the movie and reproached the robot for its advices. Scenario B: participants liked the movie and thanked the robot. In the both cases, participants had to observe three possible behaviors and to indicate if the robot provided a credible and sincere answer. Concerning the A scenario, the robot said: "I'm really sorry. This movie has good critics. I thought you liked it." Concerning the B scenario, it said:"Thus, you liked the movie! I'm happy to give you good advices.". In both cases, the three suggested behaviors were similar. The first behavior (*Apologize* on Fig. 1) was the apologize gesture. The second behavior (*showed neutral* on Fig. 1) was a neutral expression, slightly negative (the robot gently lowered its head). The third behavior (Happiness on Fig. 1) was an intense joy. The real neutral expression on Fig. 1 shows the robot basic positioning, considered like its neutral positioning. The A scenario congruent behavior was the first behavior. The B scenario congruent behavior was the third behavior.

### III. Data collection and analysis

**Questionnaire**: After each behavior, participants had to answer a two-question questionnaire: does the robot seem express what is thinks? (sincerity). Does its expression seem plausible? (credibility). Available answers were: *not at all*, *rather not*, *rather*, *totally*.

**Statistical analysis**: Data analysis used Minitab 15© software. The Khi-Deux one sample test was used to find significant answers. The accepted P level was 0.05. Data collected were sincerity and credibility points of view concerning the six behaviors (a total of three in the A scenario and a total of three in the B scenario).

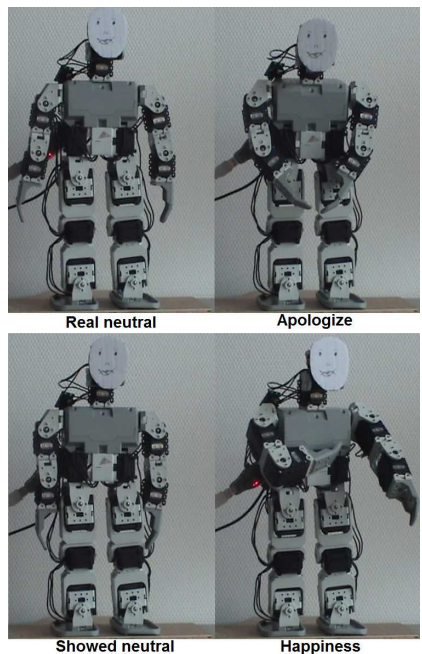

Fig. 1.  Robot key positions

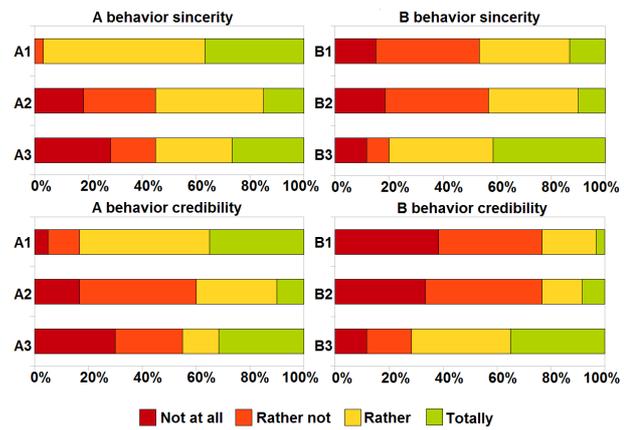

Fig. 2. Robot Credibilty and sincerity results

**Results**: In the A scenario – participants did not like the movie – participants did not make their answers random ($X^2$=189,1, df=33, p<0.001). Concerning the **first behavior**, participants significantly answered that the robot was **sincere** (60% *rather* and 36.7% *totally*, $X^2$=132.7, df=3, p<0.001; Fig. 2)  and **credible** (48.3% *rather* and 35% *totally*, $X^2$=63.0, df=3, p<0.001). Concerning the second behavior, almost half of participants found it *rather* sincere (40%, $X^2$=20.1, df=3, p<0.001; Fig. 2). However, answers were contrasted because 18.3% answered the robot was *not at all* sincere although 15% found it *totally* sincere. On the contrary, 43% of participants judged the robot *not at all* credible ($X^2$=39.7, df=3, p<0.001).

Concerning the last condition, where robot expressed the third behavior, no statistical difference appeared concerning the robot sincerity ($X^2$=5.9, df=3, p=0.117; Fig. 2). Notice a difference between gender: women had significantly more often judged the robot *not at all* sincere compared with men (41,2% of women and 23,3% of men; $X^2$=10.5, df=3, p=0.015). Concerning the credibility, participants significantly gave contrasted answers ($X^2$=12.4, df=3, p=0.006). Indeed, 30% of them judged the behavior not al all credible and 31.67% judged it *totally* credible. A minority of people expressed intermediate situations (25% and 13.3%).

In the B scenario – participants liked the movie – participants did not make their answers random ($X^2$=88.4, df=15, p<0.001). Concerning the first behavior, 38.33% of participants answered that the robot was *rather not* sincere and 33.33% *rather* sincere ($X^2$=26, df=3, p<0.001; Fig. 2). Answers about credibility are more punctuated: only 3.3% found the robot *totally* credible whereas the majority found it *rather not* or *not at all* credible (43,3% and 33,3% respectively, $X^2$=42.7, df=3, p<0.001; Fig. 2).

Concerning the second behavior, 38/3% of participants answered the robot was *rather not* sincere and 33.3% *rather* sincere ($X^2$=29.9, df=3, p<0.001; Fig. 2). Answers about credibility are more punctuated too: only 8.3% found it totally credible whereas the majority found it rather not or not at all credible (43.3% and 33.3% respectively, $X^2$=42.7, df=3, p<0.001; Fig.2). Concerning the third behavior, participants significantly answered that the robot was sincere (38.3% *rather* and 41.7% *totally*, $X^2$=56.5, df=3, p<0.001; Fig. 2) and credible (36,7% *rather* and 35% *totally*, $X^2$=26.2, df=3, p<0.001; Fig. 2).

### B. Discussion

Concerning A scenario – robot apologized for the wrong advice – the robot was judged credible and sincere when it expressed the first behavior.

This situation is the congruent one and is understood by a large sample group. Concerning both others behaviors, answers were contrasted. No one seem capable to give their opinion. The uncertain answers may reflect hesitation due to a misunderstanding of the situation. For example, concerning the third behavior credibility, participants answered around 25% to each possible answer. It may show that they make their answers random because they did not know how to think about the behavior. Thus, the second and third behaviors seem difficult to measure. The robot intention is not obvious.

Concerning the B scenario – robot congratulated itself for the good advice – participants judged that the third behavior was credible and sincere. They chose the congruent situation. Both others behaviors were judged not credible whereas sincerity seemed difficult to estimate. Actually, sincerity seem more difficult to tackle in an uncertain context than credibility.

Generally, participants judged more easily that the robot expressed what it thought even if its expression did not seem plausible. In both scenarios, congruent situation have been chosen, without ambiguity. Concerning non congruent behaviors, there were no hesitation in the case of positive emotion. On the contrary, in the case of negative emotion, the robot has been judged rather sincere, even if it was not totally credible and if it expressed positive gesture. Moreover, the second behavior, which was a negative neutral, was not chosen. However, the robot lowered its head and did a small hand movement, keeping its arms along its legs. This behavior is widely used by humans whereas the third behavior was exaggerated. Thus, since the exaggerated behavior has been chosen, and since the sincerity is the same for a neutral or positive behavior, it seems important to emphasize a gesture in the case of negative emotion to make the robot more credible and sincere.

To conclude, it is important that gesture and speech are in adequacy. By this way, humans judge a robot more credible and sincere.

## IV. Study 2: A robot acceptability despite some communication incoherence

This experiment allowed to measure a robot acceptability during an interaction with a human. Each participant had to observe a human (the experimenter) and a robot which were speaking together. There were four different dialogs. The robot randomly expressed an incoherent gesture during the experiment. The purpose was to study whether these imperfections had an impact on human perception.

Participants were 26 French students (9 women and 17 men) between 18 and 28 years old (mean: 22.5). Participants did not have the same knowledge in robotics. They did not have the same specialty (computing, mathematics, statistics) and dis not have the same level (two years after A-level and four years after A-level). The purpose of these differences was to obtain answers which were not influenced by a common environment.

### A. Experimental design

The experiment was made in an isolated room with a robot and an experimenter. During the interaction, participant has to observe the scene and to complete a questionnaire.

#### I.    Equipment
A small humanoid robot (see Fig. 1) was placed on a table, face to the experimenter. Participants sat down next to the experimenter.

#### II.    Experimental setting and data recording
First of all, the experimenter introduced the robot to participants and the context of human-robot interaction. Participants had to read the questionnaire before experiment started. When experiment started, the four dialogs were played without any break. Participant had to answer questions during the interaction. No break allowed to stay concentrated on the interaction context.

Experiment dialogs were the following:
**Dialog 1**
  *Human greetings*
  *Robot greetings*
  Human: "Help me to plan my holidays please!"

Robot: "OK, I need information about your willing."
*Human gives information about the holidays he wishes*
*Robot thanks human and* says "OK, I'm doing it!"
Human: "Thank you, goodbye."

**Dialog 2**
  Robot: " I found a campground for you."
  Human: "Thank you, it looks great!"
  *Robot is happy.*
  Robot: "Can I suggest you hiking?"
  *Human acquiesces.*
  Robot: "What kind of hiking?"
  Human: " I don't have time to think about that, make it yourself!"
  *Robot complains.*

**Dialog 3**
  Human: "Today I went hiking. It was great!"
  *Robot is happy.*
  Human: "There were a lot of garbage in the natur."
  *Robot expresses discontent*
  Human: "I collected everything in a bag to throw them away!"
  *Robot compliments human*

**Dialog 4**
  Human: "It was raining today for my hiking."
  *Robot is sad.*
  Human: "you might have given information about the weather."
  *Robot apologizes.*
  Human: "It's your fault, I don't want you to apologize!"
  Robot: "You didn't ask me to give information about the weather"
  Human: "Finally, I left the campground without paying."
  *Robot reprimands human*

#### III.    Data collection and analysis
**Questionnaire**: The questionnaire asked questions about each robot action independently, about each dialog (general questions) and about the entire interaction.

After each sentence, participants had to indicate whether they agree the following proposals:
  • The gesture is in adequacy with the sentence content.
  • The speed gesture is correct.
  • The gesture seems to be flexible and natural.
  • The robot is credible when speaking.
After each dialog, participants has to indicate whether they agree the following proposals:
  • The speech sequences are correct.
  • Answers speed of the robot and the human is correct.

- The expression of the robot is correct when listening to human.
- The robot is attentive.

At the interaction end, participant had to give her/his point of view concerning the following proposal: the robot is pleasant/nice.

The possible answers were not at all, a little, a lot, no idea; except for the second proposal: Too slow, slow, normal, fast, too fast.

**Statistical analysis**: Data analysis used Minitab 15© software. The Khi-Deux one sample test was used to find significant answers. The accepted P level was 0.05. Data collected were answers given by participants.

**Results**: Gesture is significantly *a little* flexible and natural (50.51%, $X^2=318.29$, df=3, p<0,001), whereas it was significantly *a lot* in adequacy with the sentence content (62.31%, $X^2=365.03$, df=3, p<0,001; Fig. 3). Moreover, the gesture speed was significantly *normal* (66.67%, $X^2=623.32$, df=4, p<0,001; Fig. 3). The speech sequences were significantly *a lot* correct (67.31%, $X^2=69.57$, df=2, p<0,001; Fig. 3). Only 4.81% of participants found that the robot or the human did *not at all* answered at the right speed ($X^2=50.21$, df=2, p<0,001; Fig. 3).

And only 6.73% of them found significantly that robot expressions are not correct when listening to the human ($X^2=31.27$, df=2, p<0,001; Fig. 3). Concerning the robot, participants significantly judged it *a lot* credible (60%, $X^2=353.26$, df=3, p<0,001; Fig. 3) and significantly *a lot* attentive (67.31%, $X^2=64.18$, df=2, p<0,001; Fig. 3). Finally, no one significantly found the robot unpleasant ($X^2=13.04$, df=2, p=0,001; Fig.3).

### B. Discussion

The human-robot interaction has positively been perceived (see Fig. 3) by participants. Some proposals did not have good results. Some proposals did not have good results. That illustrates the adequacy errors randomly expressed by the robot. For example, its expressions have not always been judged correct when listening to human. Robot gesture seemed only a few flexible and natural. It is normal because the robot does not have enough physical capacity and cannot express rich movements. In despite of the low quality of its actions, the robot was widely judged attentive during the interaction and **no one judged it was unpleasant**.

This study showed that human perceived positively a robot, even if its nonverbal behavior is not occasionally in adequacy with its speech.

The robot do not have an object status. Indeed, if an object in not efficient to do what it has to do, human is not anymore interested and throws it away. In the robot case, human judgment seems less severe or strict. The robot seems to be a potential social partner for the human because it is accepted in spite of its "defects".

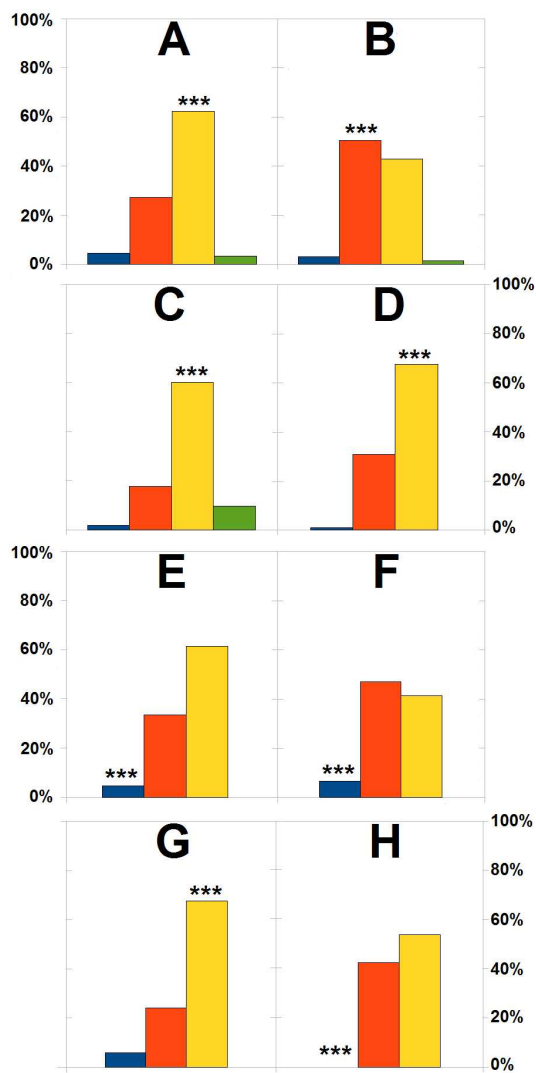

\* p<0.05, \*\* p<0.01, \*\*\* p<0.001 (Khi-Deux one sample test)

■ Not at all  ■ A little  ■ A lot  ■ No idea

Fig. 3. Robot acceptability results.
A: The gesture is in adequacy with the sentence content.
B: The gesture seems to be flexible and natural.
C: The robot is credible when speaking.
D: The speech sequences are correct.
E: Answers speed of the robot and the human is correct.
F: The expression of the robot is correct when listening to human.
F: The robot is attentive.
G: The robot is pleasant/nice.

## V. CONCLUSION

A robot nonverbal behavior has to be in adequacy with its speech in order to increase its perceived sincerity and credibility. But, in a long-term interaction context, some rare incoherence does not alter the robot credibility and sincerity. It seems that humans can "forgive" a robot to be not perfect. That indicates that humans have a kind of feeling for robots. They do not consider the robot like an ordinary object. But, consciously or not, humans personified robots. That shows that the robot can be a social partner for the human.

Moreover, we compared our first experiment results with the virtual agent results showed in [20] in order to understand the social position of robots compared to virtual agents. This comparison is interesting because virtual agents are social partner for humans and inter-act everywhere and whenever with humans. Notice that the robot results are better than virtual agents results (see Table 1). **On average, a robot seems more cred-ible and sincere than a virtual agent**. But, it is not possible to affirm that because it is possible that experi-ment has not exactly been the same. Maybe robot movement had been more explicit than agent move-ment. However, a new problematic follows from these both studies: Is a robot a better social partner than a virtual agent? Our future work will be to answer this question. We will realize a study which will compare the impact of a robot, a computer and a virtual agent on the human-machine interaction.

TABLE I
COMPARISON BETWEEN AGENT AND ROBOT SINCERITY AND CREDIBILITY

| | Apologize | | Joy | |
|---|---|---|---|---|
| | SINCERITY | CREDIBILITY | SINCERITY | CREDIBILITY |
| AGENT | 65.00%<br>Rather: 39%<br>Totally:26% | 70.00%<br>Rather: 52%<br>Totally: 18% | 74.00%<br>Rather: 52%<br>Totally: 22% | 78.00%<br>Rather: 52%<br>Totally: 26% |
| ROBOT | 97,00%<br>Rather: 60%<br>Totally: 37% | 83,00%<br>Rather: 48%<br>Totally: 35% | 80.00%<br>Rather: 38%<br>Totally: 42% | 72,00%<br>Rather: 37%<br>Totally: 35% |

## ACKNOWLEDGMENT(S)

This work has been supported by French National Research Agency (ANR) through TecSan program (project Robadom n°ANR-09-TECS-012).

Thanks to Nhung Nguyen Thuy for her help during the experiments.